\documentclass{llncs}
\usepackage[english]{babel}
\usepackage[utf8]{inputenc}
\usepackage{amsmath,amssymb,url}
\usepackage{mathabx}
\usepackage{graphicx,tikz,rotating,yh,soul}
\usepackage[all]{xy}
\xyoption{all}
\usepackage[colorinlistoftodos]{todonotes}
\usepackage[lined]{algorithm2e}

\title{INAUT, a Controlled Language for the  French Coast Pilot Books \emph{Instructions nautiques}\thanks{The final publication is available at \texttt{http://link.springer.com}.}}

\author{Yannis Haralambous \and Julie Sauvage-Vincent \and John Puentes}
\institute{Institut Mines-Télécom, Télécom Bretagne \& 
UMR~CNRS~6285 Lab-STICC\\
Technopôle Brest Iroise
CS 83818, 29238~Brest~Cedex~3, France}

\date{\today}

\begin{document}
\maketitle
\begin{abstract}
We describe INAUT, a controlled natural language dedicated to collaborative update of a knowledge base on maritime navigation and to automatic generation of coast pilot books (\emph{Instructions nautiques}) of the French National Hydrographic and Oceanographic Service SHOM. INAUT is based on French language and abundantly uses georeferenced entities.	After describing the structure of the overall system, giving details on the language and on its generation, and discussing the three major applications of INAUT (document production, interaction with ENCs and collaborative updates of the knowledge base), we conclude with future extensions and open problems.
\end{abstract}

\section*{Introduction}

\emph{Instructions nautiques} is the name of a nautical book series \cite{INweb} published by the French Marine Hydrographic and Oceanographic Service (SHOM). They are the French counterpart of the \emph{United States Coast Pilot} \cite{US}, published by the United States National Oceanic and Atmospheric Administration's Office of Coast Survey, and of the British {Admiralty Sailing Directions} \cite{UK} published by the United Kingdom Hydrographic Office.

These publications aim to supplement charts (both paper ones and ENCs = Electronic Nautical Charts), in the sense that they provide the mariner with supplemental information not in the chart.

Information for the \emph{Instructions nautiques} is provided by survey vessels, port officers, maritime officers and mariners in general. In some cases, it may require immediate update, for example to notify a shipwreck or some important change of the navigation conditions.  

The SHOM is building a knowledge base that will cover both ENCs and nautical instructions. This knowledge base will communicate with ENCs and navigation equipment and, since updates can be frequent, the \emph{Instructions nautiques} will have to be generated on-the-fly by the knowledge base. 

To summarize, we have two constraints:
\begin{enumerate}
\item the information contained in the knowledge base, has to be easily updatable by people not necessarily proficient in the ontology formalism;
\item the \emph{Instructions nautiques}, or at least part of them, have to be automatically generated out of the knowledge base.
\end{enumerate}

To fulfill constraint~1, we have built INAUT, a controlled language based on French natural language, and dedicated to the population and update of the SHOM knowledge base. Contraint~2 is fulfilled by generation of texts in INAUT out of the knowledge base. In fact, the texts generated will be in a more ``literary'' and concise version of the language, called LitINAUT (=~Literary INAUT, \S\ref{LitINAUT}) that will bring them closer to legacy human author production.




To our knowledge, INAUT is the first maritime CNL\footnote{With the exception of Seaspeak \cite{seaspeak}, a CNL defined in 1985 by the International Maritime Lecturers Association. In 2001 it evolved into SMCP (Standard Marine Communication Phrases \cite{smcp}) which is still used today. These CNLs, dedicated to oral communication between ships, are “human-only”.}

In the following, we present our model of the \emph{Instructions nautiques} (\S\ref{model}), the SHOM knowledge base (\S\ref{kb}), the controlled language (\S\ref{cnl}) and its generation (\S\ref{nlg}), the main operations (interaction with ENCs (\S\ref{elec}) and collaborative updates of the knowledge base (\S\ref{update})) as well as future extensions and open problems (\S\ref{future}).

\section{Modelling the \emph{Instructions nautiques}}\label{model}

\begin{figure}[h]
\centering\resizebox{.7\textwidth}{!}{\includegraphics{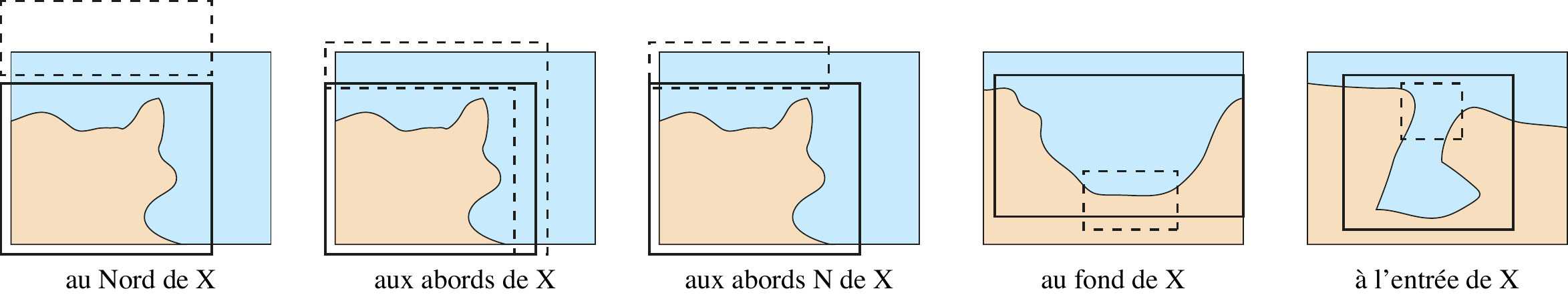}}
\caption{The main modifiers: solid polygon represents the original area $A$, dashed polygon the modified one $\mu(A)$.\label{modif}}
\end{figure}

We model the \emph{Instructions nautiques} as a set of three graphs $(\mS,\mG,\mK)$: the \emph{hierarchical structure of the document} $\mS $, the \emph{geographic areas graph} $\mG$ and the \emph{SHOM knowledge base} $\mK$ (see \S\ref{kb}). Between these graphs we have two functions: $g$ which maps some nodes of $\mS$ and of $\mK$ (those that are goereferenced) to nodes of $\mG$, and $\kappa$ that maps leaf nodes of $\mS$ to subgraphs of $\mK$. Furthermore, we have a set $\mT$ of \emph{titles of hierarchical subdivisions}, a set $\mA$ of \emph{geopolygons}, and functions $\tau$ and $\alpha$ mapping nodes of $\mS$ (resp. $\mG$) to $\mT$ (resp. $\mA$). Finally, there is a set $\mM$ of functions $\{\mu\}$ defined both as $\mT\to\mT$ and as $\mA\to\mA$, called \emph{modifiers}. Here are the details:
\begin{itemize}
\item graph $\mS $ represents the hierarchical structure of a given volume. $\mS$ is rooted, oriented and ordered. Let $V(\mS)$ be the vertices and $E(\mS)$ the edges of $\mS$;
\item the five first levels of $V(\mS)$ represent hierarchical subdivisions. Let us denote $\ell$ the level function. The root $n_0$ represents the entire document;
\item function $\tau\colon V(\mS)\to\mT$ maps every node $n$ to a title $\tau(n)$,
\item $V(\mS)$ can be written as\footnote{We denote by $\sqcup$ the disjoint union: $C=A\sqcup B\iff (C=A\cup B)\wedge(A\cap B=\emptyset)$.} $V(\mS)=V_G\sqcup V_{\neg G}$ ($V_G$ are the \emph{georeferenced nodes}) where
we have a function $g\colon V_G\to\mG$ that maps every georeferenced node to a node in $\mG$, which again is mapped to a geopolygon in $\mA$ by $\alpha$; 
\item when generating a volume of the \emph{Instruction nautiques}, the leafs of $\mS $ are mapped to subgraphs of $\mK$ through function $\kappa$. These subgraphs are then converted to text paragraphs in LitINAUT language;
\item edges $E(\mG)$ of $\mG $ represent \emph{partial inclusion} $\psubset$ in $\mA$, in the sense that we have $G'G\in E(\mG)$ (or $a(G')\psubset a(G)$) if and only if $\mathrm{Area}(a(G')\cap a(G))>0.8\,\mathrm{Area}(a(G'))$;
\item the barycenters of $\mathrm{Im}\alpha\circ g$ when restricted to $V_G\cup\ell^{-1}(i)$ for $i\in\{1,2,3\}$ follow a path on the map, which corresponds to an itinerary along the coasts of France. The extremities of this itinerary for a given volume are given in $t(n_0)$ where $n_0$ is the root of $\mS$. We call this path, the \emph{guiding path} of the volume.
\item modifiers $\mu$ serve to describe locations relatively to other locations. For example, the modifier “au nord de X” (= to the North of~X), applied to location “[cap Cerbère]” will produce “au nord de [cap Cerbère]” which is a new geographic entity, the polygon of which is calculated automatically out of $\alpha(g(\text{[cap Cerbère]}))$. Some modifiers are shown in Fig.~\ref{modif}.
\end{itemize}
\begin{figure}[t]
\resizebox{\textwidth}{!}{\includegraphics{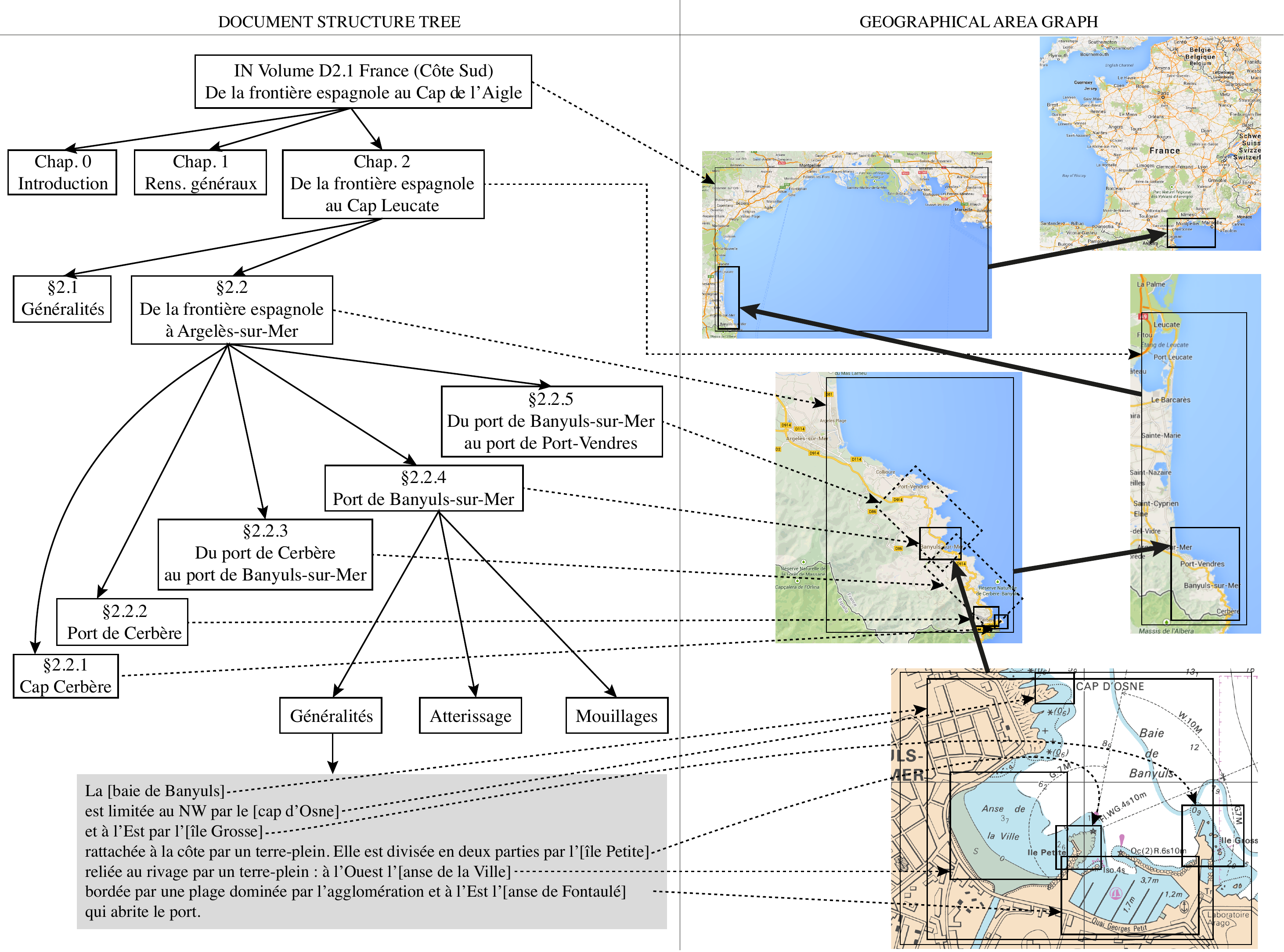}}
\caption{Document structure tree and geographic area graph for an example taken from Vol. D2.1 of the \emph{Instructions nautiques}.\label{g-d}}
\end{figure}

On Fig.~\ref{g-d} the reader can see an example of \emph{Instructions nautiques} data in our model: on the left, the document structure tree, on the right, the geographical area graph. The gray box contains the LitINAUT text generated from section \emph{Généralités} of \S2.2.4. In the text, geographical entities are marked up by brackets. Dashed arrows between the two graphs represent function $g$.

\section{The SHOM knowledge base}\label{kb}

Let us define (extending Cimiano \cite{Cim}) a knowledge base $\mK$ as being a 16-tuple
$$(\fC,\leq_C,\fA,\fSR,\fR,\fT,\fI,\fV,L_\fC,L_\fA,L_\fSR,L_\fR,L_\fI,\Sig,\Inst,\Lex)$$
where $\fC,\fA,\fSR,\fR,\fT,\fI,\fV$ are sets of concepts, attributes, simple relations, complex relations, types, instances and values, $\leq_C$ is a hierarchy of concepts, $L_\fC,L_\fA,L_\fSR$, $L_\fR,L_\fI$ are sets of names of concepts, attributes, simple relations, relations and instances, and $\Sig,\Inst,\Lex$ denote signature, instantiation, lexicalization, as follows:
\begin{enumerate}
\item $\Inst\colon\fC\to2^\fI$;
\item
the signature of an attribute is $\Sig\colon\fA\to\fC\times\fT$ and its instances $\Inst\colon\fA\to2^{\fI\times\fV}$;
\item simple relations are relations between exactly two instances, without relation attributes. Hence we have
$\Sig\colon\fSR\to\fC\times\fC$ and $\Inst\colon\fSR\to2^{\fI\times\fI}$;
\item complex relations are relations between $n$ instances ($n\geq2$) which can also have relation attributes. Hence we have:
$\Sig\colon\fR\to\bigtimes^n\fC\times\bigtimes^m\fT$ (where $\bigtimes^n$ denotes $n$-fold product) and
$\Inst\colon\mR\to2^{\bigtimes^n\fI \times\bigtimes^m\fV}$, with $n\geq2$, $m\geq0$;
\item an noteworthy difference between complex relations and simple relations, is lexicalization. Indeed, we have:
$\Lex\colon\fC\to L_\fC$, $\Lex\colon\fI\to L_\fI$, $\Lex\colon\fA\to L_\fA$, $\Lex\colon\fSR\to L_\fSR$ as expected, but $\Lex\colon\fR\to L_\fR\times\bigtimes^n L_\fSR\times\bigtimes^m L_\fA$, i.e., a relation has its own name, but requires also names for all instances involved in the relation as well as all relation attributes.
\end{enumerate}


\medskip

The concepts $\fC$ of the SHOM knowledge base $\mK$, belong to the domain of maritime navigation: ports, capes, sea currents, ships, etc.

As for $\mS$ nodes, instances $\fI$ are of two types $\fI=\fI_G\sqcup\fI_{\neg G}$: $\fI_G$ are georeferenced entities: “[baie de Banyuls]”, “[cap d'Osne]”, etc., in the sense that there is a map $g$ between $\mI$ and the graph $\mG$; $\fI_{\neg G}$ are non-georeferenced instances, such as “agglomération”, “port,” etc. They don't need to be located on the map, and their purpose is purely descriptive of the environment.

Notice that the names of $\fI_G$ instances often contain a hint to the predominant concept to which they belong (“baie”, “cap”, “port”, etc.), while in the case of $\fI_{\neg G}$ instances, their names are often names of predominant concepts \emph{per se}.


Simple relations $\fSR$ represent verbs in passive or active voice “est abrité par”, “est possible”, etc.
Notice that most relations representing a passive verb have a symmetric relation representing the corresponding active verb: “A est abrité par B” has the symmetric relation “B abrite A”;

Complex relations are $n$-ary ($n\geq2$) and can have attributes: for example “est limité par” has attribute “direction.” In this case, lexicalization requires names for all instances or attributes participating in the relation. In the case of “est limité par” the members of the relation are instances “limitant”, “limité” and attribute “à.”


\begin{figure}[t]\centering
\resizebox{.75\textwidth}{!}{\includegraphics{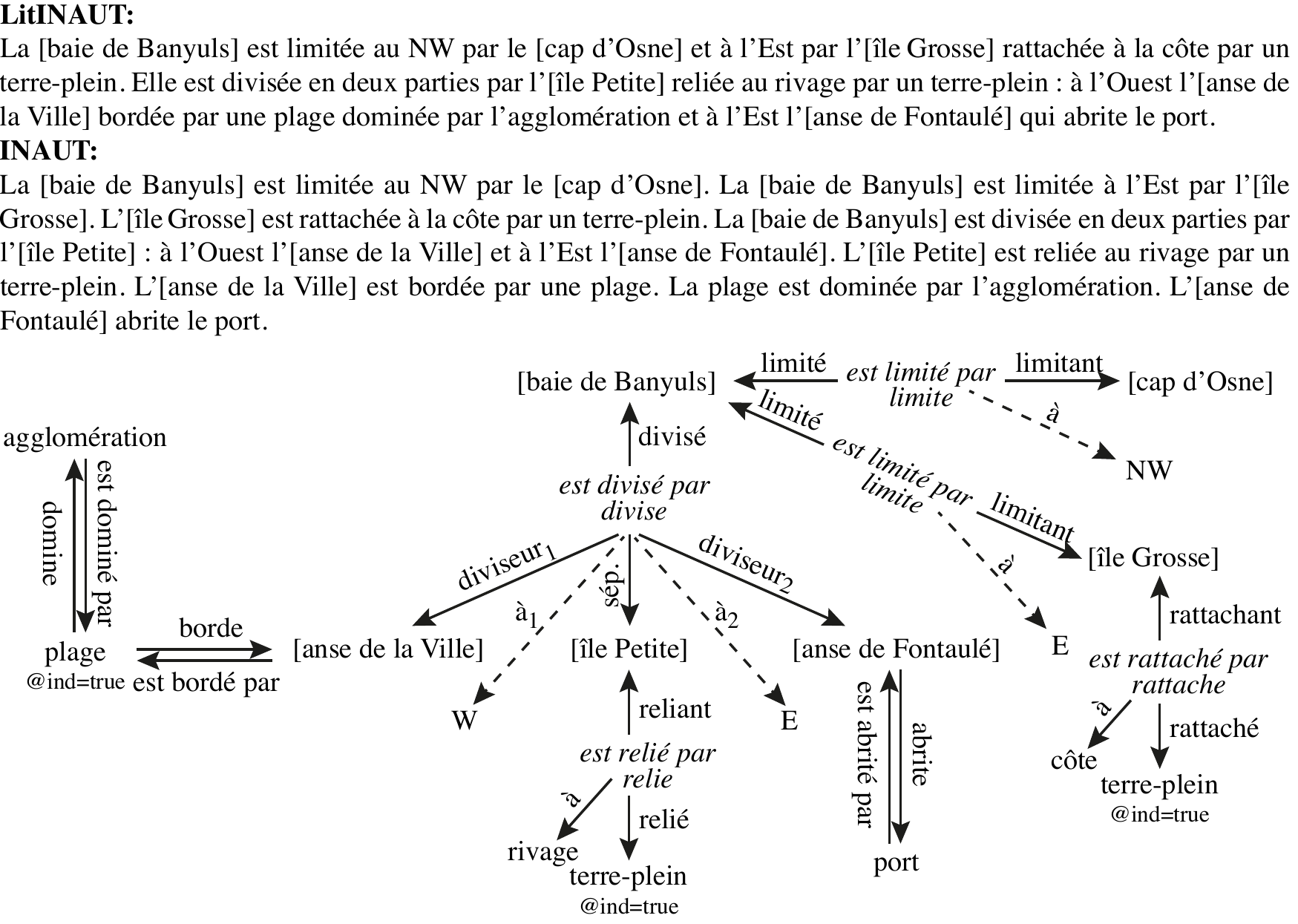}}
\caption{A paragraph in INAUT, LitINAUT, and represented in the knowledge base.\label{kb}}
\end{figure}

As an example, the reader can see on Fig.~\ref{kb} the sentence of Fig.~\ref{g-d}, represented in LitINAUT, INAUT and as a subgraph of $\mK$. Instances “[baie de Banyuls]”, “[cap d'Osne]”, etc. belong to $\fI_G$. Instances “côte”, “plage”, “port”, “rivage”, belong to $\fI_{\neg G}$. Complex relations are reified as nodes. Attributes of instances have been included underneath, marked by character @. 


\section{The controlled languages INAUT and LitINAUT}\label{cnl}

INAUT is a controlled language with a rather large vocabulary (based on the existing \emph{Instructions nautiques} corpus) but with a simple syntax, given by the following grammar:
\begin{quote}
S $\to$ NP VP\\
NP $\to$ \textsc{modif det} NN $|$ \textsc{det} NN $|$ NN\\
NN $\to$ \textsc{adj} NN $|$ NN \textsc{adj} \textsc{noun}\\
VP $\to$ \textsc{verb} NP $|$ \textsc{verb} NP PP\\
PP $\to$ \textsc{prep} \textsc{det} NN $|$ \textsc{prep} NN
\end{quote}
where symbols in small caps are terminal, all \textsc{noun}s belong to $L_\fI$ (the set of lexical references for instances of $\mK$) and to $\fV$ (the set of values of attributes of $\mK$), all \textsc{verb}s belong to $L_\fSR\cup L_\fR$, all \textsc{adj}ectives belong to $\fV$, and \textsc{modif}iers, \textsc{det}erminants and \textsc{prep}ositions belong to a closed list.

The \textsc{verb}, always in 3rd person or in the infinitive, can be active or passive. In most cases it is possible to change the voice of the verb, which implies a permutation of the NPs in \textsc{subject} and \textsc{object} position, leaving the PPs intact:
\begin{quote}
La [baie de Banyuls] est limitée par le [cap d'Osne] au NW.\\
Le [cap d'Osne] limite la [baie de Banyuls] au NW.
\end{quote}

Definite articles are used for all instances in $\mG$ the names of which start with the name of a concept to which belongs the instance: for example, the name “baie de Banyuls” starts with “baie” (=~bay) which is the name of a concept in~$\fC$, hence in INAUT the definite article is used: “la [baie de Banyuls]”.

Otherwise, no article is used:
\begin{quote}
[Notre-Dame de la Salette] est un amer remarquable à l'WSW du port.
\end{quote}

Instances in $\fI_{\neg G}$ are, by default, used with definite articles. When an indefinite article is required, the information is stored in a dedicated attribute. Indefinite articles are used in \textsc{object} position only:

\begin{quote}
L'[anse de la Ville] est bordée par \ul{une} plage. \ul{La} plage est dominée par l'agglomération.
\end{quote}

Modifiers are represented by (a closed set of) expressions outside the brackets of the geographic entity: in “au fond de l'[anse de la Ville]”, we have modifier “au fond de” and entity “[anse de la Ville].” In $\mK$ there is a \emph{modifier relation} whenever a modifier is used. This relation does not produce INAUT text but serves to connect subgraphs in $\mK$ during content determination (\S\ref{c-d}).


\medskip

In the following sections we will discuss the three main operations of controlled languages INAUT and LitINAUT: generation (\S\,\ref{nlg}), interaction with ENCs (\S\,\ref{elec}), collaborative updates of the knowledge base (\S\,\ref{update}).

\section{Controlled language generation}\label{nlg}

One of the design goals of our system is to be able to produce automatically a large part of the \emph{Instructions nautiques}, so that after collaborative updates new versions of the entire document can automatically be produced.

We have divided the task into two stages: 
(1) produce INAUT text corresponding to a given leaf node of $\mS$;
(2) convert INAUT language into LitINAUT.


Suppose given a leaf node $S$ of $\mS $. Producing the corresponding INAUT text is typically a Natural Language Generation problem.

Reiter \& Dale \cite[\S\,3.3]{nlg} divide the language generation task into seven subtasks: content determination, document structuring, lexicalisation, aggregation, referring expression generation, linguistic realisation and structure realisation.

\paragraph{Content determination.}\label{c-d}

\begin{algorithm}[tb]\fontsize{8pt}{8pt}\selectfont
\KwIn{$\mS,\mG,\mK$ and a leaf node $S\in\mS$}
\KwResult{The subgraph $K\subset\mK$ which represents the text corresponding to $S$}
$G_S\leftarrow g(S)$\;
$K\leftarrow\emptyset$\;
\For{$G\in\mG$}{
\If{$G\psubset G_S$ and $g^{-1}(G)\in\fI_G$}{
$K\leftarrow K\cup g^{-1}(G)$\;
}
}
\For{$k\in K$}{
\If{$\exists k'$ such that $kk'\in\mathrm{Undirected}(\mK)$ and $kk'\not\in K$}
{\If{$k'\in\fI_{\neg G}$ or $k'\in\fV$ or $(k'\in\fI_G$ and $g(k')\psubset G_S)$}
{
$K\leftarrow K\cup kk'$\;
}
\If{$k'\in\fR$ and $\exists k''$ member of $k'$ such that $g(k'')\psubset G_S$}{
$K\leftarrow K\cup kk'$\;
\For{$m$ member of $k'$}{
$K\leftarrow K\cup k'm$\;
}
}
}
}
\medskip

\caption{Content determination algorithm.}\label{alg:algo}
\end{algorithm}

Using Algorithm~\ref{alg:algo}, we find the subgraph $K$ of~$\mK$ which is geographically the most relevant to $S$. We apply tags to its connected components using a rule-based decision system: for example, when a connected component contains the instance ``mouillage'' then it is tagged as belonging to a leaf node of type “Mouillages.” If after applying the rules no tag has been affected, then the component belongs to a leaf node of default type “Généralités.”

\paragraph{Document structuring.}
This is the most difficult phase since it deals with the order in which sentences are written. Let $K$ bet the subgraph of $\mK$ to be converted into INAUT.

We subdivide the task in four subtasks:
\begin{enumerate}
\item sort connected components $K_i$ of $K$;
\item for each component find a starting node $s$;
\item find the order in which the relations of each component will be converted into INAUT, starting from $s$;
\item convert relations into INAUT in the order given by 1 and 3.
\end{enumerate}
For subtask~1, we will sort components. The sorting criteria are: (a) if there is a significant difference in size between the cumulated geographic areas of two components, the larger one will precede the smaller one, (b) otherwise, calculate the barycenters of cumulated geographic areas of components; the path defined by their barycenters should be roughly parallel to the guiding path of the volume. For example, on Fig.~\ref{g-d} the areas of nodes \S2.2.1--\S2.2.5 follow a SE to NW direction, this direction can be chosen for the order of connected components.

To accomplish steps~2 and 3 we define weights $w$ on nodes and relations. Calculation of these weights is based on criteria we will describe below, as well as on training using machine learning algorithms on the existing \emph{Instructions nautiques} corpus.

Notice that we use undirected graphs since every edge can be inverted by changing voice. 


The first and most obvious criterion is the relation between nodes in $K_i$ and the parent of the leaf node of $\mS$ that established the connection with $\mK$ (i.e., $\kappa^{-1}(K)$). If among the nodes there is one whose geographic area and/or name matches as closely as possible the one of the parent of the leaf node, it is a good choice. For example, in our case, node \S2.2.4 of $\mS$ is “Port de Banyuls-sur-Mer” which is much closer to “baie de Banyuls” than to “cap d'Osne”, both in terms of geographic area than simply of string comparison of names.

The second criterion for choosing the starting node is its position in the $\mG$ lattice. Let $k_m = g^{-1}(\max_{\mG} g(K_i))$. If $k_m\in K_i$ then it is an obvious choice. Otherwise we take local maxima in $\mG$ and proceed with weighting. 

Finally, another criterion is of semantic nature, the one of “interest” for the navigator: an order can be established between concepts to which instances of $K_i$ belong, for example a port instance will be more interesting than a beach instance. This weight is inherited by neighboring nodes: for instance, a bay containing a port is more interesting than a bay containing a beach, etc.

The “semantic weight” of instances can be calculated by machine learning.

Once the starting point has been established, we proceed to subtask~3. We will use a variant of DFS (depth-first search) to search $K_i$.

Subtask~4 is the simplest one: from relations in $\mK$ we build INAUT sentences, by applying rules, for example choosing the verb's voice according to the direction of the search in $K_i$, adding articles matching nouns, etc. There still remains a difficulty: finding the right order of prepositional phrases, as in:
\begin{quote}\small
La [baie de Banyuls] est limitée par le [cap d'Osne] au NW.\\
La [baie de Banyuls] est limitée au NW par le [cap d'Osne].
\end{quote}
When the difference may be purely stylistic (as above), the order can be obtained by machine learning. In other cases, such as in “est divisé par” of Fig.~\ref{g-d}, it is mandatory to group some relations: the text representing “à$_i$” must immediately follow the one representing “diviseur$_i$” since indices disappear in the textual realization and only proximity allows to distinguish the divisors of the entity.

Another important phenomenon is text added by default: for example, to realize relation “est divisé par” we need to add the number of divisors, this is done by counting the members of the relation of type “diviseur” and generate “en deux parties” (=~in two parts).



\paragraph{Aggregation and referring expressions generation: LitINAUT language.}\label{LitINAUT}

At this stage, generation of INAUT has been completed. The result, as it can be seen in Fig.~\ref{kb}, is not very eloquent, but remains closely related to the structure of $\mK$, so that it is easier for contributors to supply modifications and additions written in INAUT. To produce a human readable text as part of automatically generated \emph{Instructions nautiques} document, we need two extra steps: aggregation of several sentences into a single one, and generation or referring expressions. The result of these two operations is called LitINAUT language.

By the fact of using DFS to search $K_i$, often the object of a sentence is the subject of the following one. Aggregation merges them into a single sentence:
\begin{quote}\small
L'[anse de la Ville] est bordée par une plage.\\
La plage est dominée par l'agglomération.\\
$\to$ L'[anse de la Ville] est bordée par une plage, dominée par l'agglomération.
\end{quote}
In other cases, consecutive sentences have the same object and the same verb; in that case we use conjunction:
\begin{quote}\small
La [baie de Banyuls] est limitée au NW par le [cap d'Osne].\\
La [baie de Banyuls] est limitée à l'Est par l'[île Grosse].\\
$\to$ La [baie de Banyuls] est limitée au NW par le [cap d'Osne] et à l'Est par l'[île Grosse].
\end{quote}
When we have object identity but with different verbs, referring expressions are generated:
\begin{quote}\small
La [baie de Banyuls] est divisée en deux parties par\ldots
$\to$ Elle est divisée en deux parties par\ldots
\end{quote}
In some cases, text is omitted from the realization because it is obtained from the context: for example, in realizing the text corresponding to a subdivision of type “Mouillages” (=~mooring), we will systematically omit the part “Le mouillage est autorisé à” since it is implied by the subdivision title.

These are just some examples of mechanisms used to convert INAUT into LitINAUT. Work is in progress to enhance the result and bring it closer to legacy (human authored) text.

\section{Interaction with ENCs}\label{elec}

As said in the introduction, \emph{Instructions nautiques} are defined as a complement to charts, and, in particular, to ENCs. Therefore it is important to define interactions between $\mK$ and ENCs, via INAUT. By specifying, for example, an area of interest on an ENC (for example, by drawing a zone on a touchscreen) the user may receive LitINAUT text in return. Generating this text automatically has the advantage of being (a) limited to the zone of interest given by the user; (b) conform to local conditions, for example time of the day (some relations or attributes in $\mK$ may be time-dependent) or meteorological conditions, or parameters of the user's vessel (size, tonnage, etc.); (c) up-to-date, since other users may constantly provide new information.

To provide adaptive LitINAUT text, we first position the area $U$ given by the user in $\mA$, and hence in $\mG$. Knowing the subgraph of $\mG$ that matches as closely as possible $U$ in $\mA$, we find the relevant nodes in $\mK$ by going through $g^{-1}$. These nodes form a subgraph of $\mK$ and we apply the techniques described in~\ref{nlg} to generate the corresponding text in LitINAUT.

Additional structure can be added to the text sent to the ENC device, so that the user can filter the text and display only specific types of information, as for example information on mooring, landing, etc.

\section{Collaborative updates of the knowledge base}\label{update}

It is important for the SHOM knowledge base to be kept constantly up-to-date. To achieve this goal, INAUT will be used as a tool for collaborative update. Indeed, INAUT has been designed as the optimal compromise between easiness of use (since contributors have a priori no KM proficiency) and formality (as the knowledge base will be fed directly by the incoming data).

To make the system more robust, we validate on two levels. First, the lexical and syntactic level: a Web interface analyzes segments written in INAUT and validates them. In case of errors it provides correction hints. Second,
the semantic level: a human controller monitors incoming INAUT data which, depending on the contributors trust level are automatically fed into the knowledge base (with the possibility of making the modification retroactive) or are stored in a waiting list until manual validation.

\section{Conclusion and future work}\label{future}

We have described the controlled natural language INAUT (and its variant LitINAUT) which is used for the update of the SHOM maritime knowledge base, for automatic generation of \emph{Instructions nautiques} documents and for interaction with ENCs. 

Among our plans is the extension of INAUT into a QA system. This requires extension of INAUT to interrogative sentences and increased use of the concept hierarchy.
Another extension deals with the issue of \emph{dangerosity}. Indeed, one of goals of \emph{Instructions nautiques} is to alert the navigator on possible dangers. 
Ideally, the ENC should automatically send queries about dangerosity to the knowledge base involving the current position of the vessel and various external conditions, and in case of a positive answer, alert the navigator by all means possible. Special NLG techniques can then be used, since the communicative goal will not be simply to inform, but to alert.



\bibliographystyle{splncs03}
\bibliography{cnl2014}

\end{document}